\documentclass[runningheads]{llncs}

\usepackage[T1]{fontenc}
\usepackage{amsmath}
\usepackage{upgreek}
\usepackage{multirow}
\usepackage{multicol}
\usepackage{comment}
\usepackage{tabularx}
\usepackage{booktabs}
\usepackage{arydshln}
\usepackage{siunitx}

\usepackage{hyperref}
\usepackage{url}
\usepackage{graphicx}

\usepackage[dvipsnames]{xcolor}

\newcolumntype{H}{>{\setbox0=\hbox\bgroup}c<{\egroup}@{}}
\newcolumntype{P}[1]{>{\centering\arraybackslash}p{#1}}
\newcolumntype{R}{>{\raggedleft\arraybackslash}X}
\newcolumntype{C}{>{\centering\arraybackslash}X}
\newcommand\Tstrut{\rule{0pt}{2.6ex}}
\newcommand\Bstrut{\rule[-0.9ex]{0pt}{0pt}}

\usepackage{environ}

\newcommand{\compactparagraph}[1]{%
\medskip\noindent\textbf{#1}\hspace{0.2cm}
}

\NewEnviron{eqblock}[1]{%
    \vspace{0pt} 
    \scalebox{0.9}{
    \begin{minipage}[t]{0.5\linewidth}
        \vspace{0pt} 
        \centering
        \textbf{#1}
        \vspace{-0.1cm}\BODY
    \end{minipage}
    }
}

\newcommand{\ra}[1]{\renewcommand{\arraystretch}{#1}}

\begin{document}
\title{Minimal Convolutional RNNs Accelerate Spatiotemporal Learning}

\author{Coşku Can Horuz\inst{1} \and
Sebastian Otte\inst{1} \and
Martin V. Butz\inst{2} \and \\
Matthias Karlbauer\inst{2}
}

\authorrunning{Horuz et al.}

\institute{Adaptive AI Research Group, University of Lübeck, Germany\\
\and
Neuro-Cognitive Modeling Group, University of Tübingen, Germany}
\maketitle

\begin{abstract}
We introduce MinConvLSTM and MinConvGRU, two novel spatiotemporal models that combine the spatial inductive biases of convolutional recurrent networks with the training efficiency of minimal, parallelizable RNNs. 
Our approach extends the log-domain prefix-sum formulation of MinLSTM and MinGRU to convolutional architectures, enabling fully parallel training while retaining localized spatial modeling. 
This eliminates the need for sequential hidden state updates during teacher forcing---a major bottleneck in conventional ConvRNN models.
In addition, we incorporate an exponential gating mechanism inspired by the xLSTM architecture into the MinConvLSTM,  which further simplifies the log-domain computation.
Our models are structurally minimal and computationally efficient, with reduced parameter count and improved scalability.
We evaluate our models on two spatiotemporal forecasting tasks: Navier-Stokes dynamics and real-world geopotential data.
In terms of training speed, our architectures significantly outperform standard ConvLSTMs and ConvGRUs.
Moreover, our models also achieve lower prediction errors in both domains, even in closed-loop autoregressive mode.
These findings demonstrate that minimal recurrent structures, when combined with convolutional input aggregation, offer a compelling and efficient alternative for spatiotemporal sequence modeling, bridging the gap between recurrent simplicity and spatial complexity.

\keywords{RNN \and ConvLSTM \and ConvGRU \and MinConvLSTM \and MinConvGRU \and Geopotential \and Navier-Stokes.}
\end{abstract}

\section{Introduction}
Spatiotemporal sequence forecasting has long been a critical challenge across a variety of applications such as video prediction \cite{shi2017deep} and weather nowcasting \cite{ravuri2021skilful}. These tasks typically involve predicting a sequence of high-dimensional spatial observations conditioned on a history of previous frames. In particular, for scenarios like precipitation nowcasting, where timely and localized predictions are crucial, modeling both spatial and temporal correlations efficiently and accurately is essential \cite{finn2016unsupervised}.

Previously, convolutional variants of recurrent neural networks, most notably the Convolutional Long Short-Term Memory (ConvLSTM) network \cite{shi2015convolutional}, along with similar approaches \cite{Ballas2016Delving} have proven effective for such tasks. By integrating convolutional operations into the input-to-state and state-to-state transitions, the ConvLSTM captures local spatial dependencies while modeling temporal evolution through recurrence. Stacked in encoder-forecaster architectures, these models have demonstrated strong performance on various benchmarks \cite{shi2015convolutional,karlbauer2024comparing}. However, their inherently sequential nature poses significant limitations in training efficiency, especially as the prediction horizon increases.

In parallel, the broader sequence learning community has begun to question the scalability of existing paradigms (particularly Transformers) fueling renewed interest in classical recurrent models and inspiring new architectures that emphasize parallelization, efficiency, and simplicity. In this context, heavily simplified architectures such as MinLSTM and MinGRU \cite{feng2024were} remove hidden state dependencies from gating mechanisms, enabling fully parallel training via prefix-sum, also known as the parallel scan algorithm \cite{blelloch1990prefix}. Rather than increasing architectural complexity, these models demonstrate that effective sequence modeling can be achieved through minimal, well-structured recurrence, a principle similarly embraced by a variety of modern RNN approaches \cite{gu2022efficiently,smith2023simplified,Orvieto2023Resurrecting}. 

In this paper, we transfer the efficiency of linear and scan-compatible recurrent architectures to the spatiotemporal domain by proposing minimalized convolutional RNNs presented in two flavors: MinConvLSTM and MinConvGRU. These models extend the log-domain scan-based formulation of MinRNNs
to the ConvRNN framework, eliminating the need for sequential hidden state updates during teacher forcing, while preserving local spatial sensitivity through its convolutional structure.
Additionally, we adapt the idea of exponential gating from \cite{Beck2024xLSTM} to replace the traditional sigmoidal gates in our MinConvLSTM architecture, further simplifying the model's computation in logarithmic space.

Empirical results across two benchmarks (fluid and geopotential dynamics) show that our models significantly accelerate training over conventional ConvRNNs (by a factor of 3 to 5), while achieving improved prediction accuracy in both domains, even in closed-loop autoregressive mode. These findings suggest that minimal, parallel recurrent structures offer a compelling direction for future spatiotemporal sequence modeling.

\section{Foundations}

In this section, we revisit the relevant key developments in recurrent sequence modeling, namely gated RNNs, their convolutional extensions, as well as recent work on linear, scan-compatible architectures.

\subsection{Classical Gated Memory Units} 
The decades-old long short-term memory (LSTM) \cite{hochreiter1997long} and gated recurrent unit (GRU) \cite{cho2014learning} remain backbones of many modern architectures and are applied across a wide range of domains \cite{Beck2024xLSTM,feng2024were,greff2015lstm}. These models introduced gating mechanisms to regulate information flow, effectively addressing the vanishing and exploding gradient problem. 
Their design incorporates controlled nonlinearities: sigmoidal gates that modulate the hidden state dynamics. 
Notably, these gates themselves depend on the previous hidden states, thereby introducing nonlinear dependencies into the state evolution. 
This gating not only stabilizes state propagation across time steps but also enables the model to track and maintain relevant temporal context, referred to as (implicit) state tracking \cite{chung2014empirical,greff2015lstm,merrill2024illusion}.

\begin{eqblock}{LSTM}
\begin{align*}
    \boldsymbol{\upphi}^{t} &= \sigma\left( \operatorname{Linear}_{\upphi}[\mathbf{x^t}, \mathbf{h^{t{-}1}}] \right)\\
    \boldsymbol{\upiota}^{t} &= \sigma\left( \operatorname{Linear}_{\upiota}[\mathbf{x^t}, \mathbf{h^{t{-}1}}] \right)\\
    \tilde{\mathbf{s}}^{t} &= \tanh\left( \operatorname{Linear}_{s}[\mathbf{x^t}, \mathbf{h^{t{-}1}}] \right)\\
    \mathbf{s}^{t} &= \boldsymbol{\upphi}^{t} \odot \mathbf{s}^{t{-}1} + 
    \boldsymbol{\upiota}^{t} \odot \tilde{\mathbf{s}}^{t}\\
    \boldsymbol{\upomega}^{t} &= \sigma\left( \operatorname{Linear}_{\upomega}[\mathbf{x^t}, \mathbf{h^{t{-}1}}] \right)\\
    \mathbf{h}^{t} &= \boldsymbol{\upomega}^{t} \odot \tanh\left(\mathbf{s}^{t} \right)
\end{align*}
\end{eqblock}
\begin{eqblock}{GRU}
\begin{align*}
    \mathbf{z}^{t} &= \sigma\left( \operatorname{Linear}_{z}[\mathbf{x^t}, \mathbf{h^{t{-}1}}] \right)\\
    \mathbf{r}^{t} &= \sigma\left( \operatorname{Linear}_{r}[\mathbf{x^t}, \mathbf{h^{t{-}1}}] \right)\\
    \tilde{\mathbf{h}}^{t} &= \tanh\left( \operatorname{Linear}_{h}[\mathbf{x^t}, \mathbf{r}^{t} \odot {\mathbf{h}}^{t{-}1}]\right)\\
    \mathbf{h}^{t} &= \left(1 - \mathbf{z}^{t} \right) \odot \mathbf{h}^{t{-}1} + \mathbf{z}^{t} \odot \tilde{\mathbf{h}}^{t}
\end{align*}
\end{eqblock}
\medskip%
\\
In the LSTM, $\upphi, \upiota, \upomega$ are forget, input, and output gates, controlling the information flow through cell state $\mathbf{s}$ and hidden state $\mathbf{h}$. The GRU simplifies this formulation and uses only reset and update gates (i.e., $\mathbf{r}, \mathbf{z}$) with an exponential moving average (EMA) scheme to control the hidden state flow.

\subsection{Convolutional RNNs}
ConvLSTM \cite{shi2015convolutional} and ConvGRU \cite{Ballas2016Delving} enable the modeling of spatial structure through convolution and temporal dynamics via nonlinear recurrence.
This is accomplished by casting an RNN across the spatial domain as a grid of locally connected modules, each applying the same computations with shared weights, while evolving distinct hidden states driven by local receptive fields on the input and neighboring states. 
Technically, this transition is achieved by replacing the linear operations with convolution. The notation is analogous to the LSTM and GRU equations:\\
\newcommand{\novela}[1]{\textcolor{PineGreen}{#1}}
\newcommand{\novelb}[1]{\textcolor{MidnightBlue}{#1}}
\newcommand{\novelc}[1]{\textcolor{Peach}{#1}}

\vspace{-0.4cm}
\begin{eqblock}{ConvLSTM}
\begin{align*}
    \boldsymbol{\upphi}^{t} &= \sigma\left( \novela{\operatorname{Conv}_{\upphi}}[\mathbf{x^t}, \mathbf{h^{t{-}1}}] \right)\\
    \boldsymbol{\upiota}^{t} &= \sigma\left( \novela{\operatorname{Conv}_{\upiota}}[\mathbf{x^t}, \mathbf{h^{t{-}1}}] \right)\\
    \tilde{\mathbf{s}}^{t} &= \tanh\left( \novela{\operatorname{Conv}_{s}}[\mathbf{x^t}, \mathbf{h^{t{-}1}}] \right)\\
    \mathbf{s}^{t} &= \boldsymbol{\upphi}^{t} \odot \mathbf{s}^{t{-}1} + 
    \boldsymbol{\upiota}^{t} \odot \tilde{\mathbf{s}}^{t}\\
    \boldsymbol{\upomega}^{t} &= \sigma\left( \novela{\operatorname{Conv}_{\upomega}}[\mathbf{x^t}, \mathbf{h^{t{-}1}}] \right)\\
    \mathbf{h}^{t} &= \boldsymbol{\upomega}^{t} \odot \tanh\left(\mathbf{s}^{t} \right)
\end{align*}
\end{eqblock}
\begin{eqblock}{ConvGRU}
\begin{align*}
    \mathbf{z}^{t} &= \sigma\left( \novela{\operatorname{Conv}_{z}}[\mathbf{x^t}, \mathbf{h^{t{-}1}}] \right)\\
    \mathbf{r}^{t} &= \sigma\left( \novela{\operatorname{Conv}_{r}}[\mathbf{x^t}, \mathbf{h^{t{-}1}}] \right)\\
    \tilde{\mathbf{h}}^{t} &= \tanh\left( \novela{\operatorname{Conv}_{h}}[\mathbf{x^t}, \mathbf{r}^{t} \odot {\mathbf{h}}^{t{-}1}]\right)\\
    \mathbf{h}^{t} &= \left(1 - \mathbf{z}^{t} \right) \odot \mathbf{h}^{t{-}1} + \mathbf{z}^{t} \odot \tilde{\mathbf{h}}^{t}
\end{align*}
\end{eqblock}
\medskip%
\\
Although these equations use the same notation, the variable tensors obtain an extra spatial dimension. 

ConvLSTM and ConvGRU achieve competitive performance in spatiotemporal benchmarks and find application in various domains \cite{karlbauer2019distributed,ravuri2021skilful,karlbauer2024comparing}. While incorporating powerful tools, similar to nonlinear recurrent networks, they are doomed to be stuck small scaled due to their sequential-only nature.

\subsection{Linear Recurrent Models}
Recently introduced state space models (SSMs) address the scaling limitation of nonlinear recurrent models and offer a linear solution. Starting their journey with the HiPPO theory \cite{gu2020hippo}, which introduces structured weight matrices for linear recurrences, and advancing the model efficiency with the recurrence-convolution equivalence \cite{gu2021combining}, SSMs ignited the era of linear RNNs. These models, such as S$4$ \cite{gu2022efficiently}, S$5$ \cite{smith2023simplified}, S$6$ aka Mamba \cite{gu2024mamba}, and S$6.2$ aka Mamba$2$ \cite{dao2024mamba2} showed promising results in sequence modeling reaching and exceeding the performance of Transformers \cite{vaswani2017attention}. In fact, SSMs can be computed in logarithmic parallel time using a linear number of processors and require only linear memory, while Transformers exhibit logarithmic parallel time complexity using a quadratic number of processors and incur quadratic memory usage due to their global self-attention mechanism.  During autoregressive inference, SSMs (since they are RNNs) require only constant memory.

Since the advent of SSMs, several other recurrent linear models have been developed using different approaches. The xLSTM \cite{Beck2024xLSTM} combines the SSM architecture with attention (mLSTM), while also using sequential mode recurrence (sLSTM). In addition, some of the linear RNN models use the parallel scan algorithm \cite{blelloch1990prefix}, instead of exploiting the convolution-recurrence equivalence.
This approach yields logarithmic parallel time complexity and alleviates the nontrivial convolution kernel computation which cannot handle irregularly sampled time series \cite{smith2023simplified}. Particularly \cite{martin2018parallelizing} and \cite{qin2023hierarchically} exploited the properties of the parallel scan in the linear RNN context, achieving high scalability and promising performance in numerous domains.

\subsection{Minimal RNN Variants}
The recently proposed MinLSTM and MinGRU models \cite{feng2024were}, similar to earlier contributions \cite{martin2018parallelizing}, simplify classical LSTM and GRU architectures by removing nonlinear recurrent relations, thereby also reducing the number of parameters. The resulting linear state update can be computed using a parallel scan, enabling full model parallelization.\\
\begin{eqblock}{MinLSTM}
\begin{align*}
    \boldsymbol{\upphi}^{t} &= \sigma\left( \operatorname{Linear}_{\upphi}[\mathbf{x^t}] \right)\\
    \boldsymbol{\upiota}^{t} &= \sigma\left( \operatorname{Linear}_{\upiota}[\mathbf{x^t}] \right)\\
    \hat{\boldsymbol{\upphi}}{}^{t},~\hat{\boldsymbol{\upiota}}{}^{t} &= 
    \frac{\boldsymbol{\upphi}^{t}}{\boldsymbol{\upphi}^{t}  + \boldsymbol{\upiota}^{t}},
    ~\frac{\boldsymbol{\upiota}^{t}}{\boldsymbol{\upphi}^{t}  + \boldsymbol{\upiota}^{t}}
    \\
    \tilde{\mathbf{h}}^{t} &= \operatorname{Linear}_{h}[\mathbf{x^t}]\\
    \mathbf{h}^{t} &= \hat{\boldsymbol{\upphi}}{}^{t} \odot \mathbf{h}^{t{-}1} + 
    \hat{\boldsymbol{\upiota}}{}^{t} \odot \tilde{\mathbf{h}}^{t}\\
\end{align*}
\end{eqblock}
\begin{eqblock}{MinGRU}
\begin{align*}
    \mathbf{z}^{t} &= \sigma\left( \operatorname{Linear}_{z}[\mathbf{x^t}] \right)\\
    \tilde{\mathbf{h}}^{t} &= \operatorname{Linear}_{h}[\mathbf{x^t}]\\
    \mathbf{h}^{t} &= \left(1 - \mathbf{z}^{t} \right) \odot \mathbf{h}^{t{-}1} + \mathbf{z}^{t} \odot \tilde{\mathbf{h}}^{t}
\end{align*}
\end{eqblock}
\medskip%
\\
The notation mirrors that of the standard LSTM and GRU formulations. Both MinLSTM and MinGRU omit hidden state feedback in the gates, removing nonlinear recurrence and enabling parallelizable updates. In MinLSTM, the forget and input gates are normalized to form a convex combination of both gating outputs. Also the output gate is discarded. MinGRU further simplifies the GRU structure by using a single update gate (omitting the reset gate) for EMA-style candidate blending. These changes reduce parameter count and allow efficient training, implemented in log-domain where multiplications become additions and the prefix-sum algorithm can be applied (see \cite{heinsen2023parallelization} for details).

\section{Minimal Convolutional RNNs}

In this section, we extend minimal RNNs \cite{feng2024were} to the spatiotemporal domain by introducing convolutional variants, namely, MinConvLSTM and MinConvGRU. We further optimize MinConvLSTM using exponential gating.

\subsection{MinConvLSTM and MinConvGRU}
We substitute the linear input aggregation for all the gates and the input (new hidden state candidates) with convolutions and obtain the following equations:\\
\begin{eqblock}{MinConvLSTM}
\begin{align*}
    \boldsymbol{\upphi}^{t} &= \sigma\left( \novelb{\operatorname{Conv}_{\upphi}}[\mathbf{x^t}] \right)\\
    \boldsymbol{\upiota}^{t} &= \sigma\left( \novelb{\operatorname{Conv}_{\upiota}}[\mathbf{x^t}] \right)\\
    \tilde{\mathbf{h}}^{t} &= \novelb{\operatorname{Conv}_{h}}[\mathbf{x^t}]\\
    \hat{\boldsymbol{\upphi}}{}^{t},~\hat{\boldsymbol{\upiota}}{}^{t} &= 
    \frac{\boldsymbol{\upphi}^{t}}{\boldsymbol{\upphi}^{t}  + \boldsymbol{\upiota}^{t}},
    ~\frac{\boldsymbol{\upiota}^{t}}{\boldsymbol{\upphi}^{t}  + \boldsymbol{\upiota}^{t}}
    \\
    \mathbf{h}^{t} &= \hat{\boldsymbol{\upphi}}{}^{t} \odot \mathbf{h}^{t{-}1} + 
    \hat{\boldsymbol{\upiota}}{}^{t} \odot \tilde{\mathbf{h}}^{t}\\
\end{align*}
\end{eqblock}
\begin{eqblock}{MinConvGRU}
\begin{align*}
    \mathbf{z}^{t} &= \sigma\left( \novelb{\operatorname{Conv}_{z}}[\mathbf{x^t}] \right)\\
    \tilde{\mathbf{h}}^{t} &= \novelb{\operatorname{Conv}_{h}}[\mathbf{x^t}]\\
    \mathbf{h}^{t} &= \left(1-\mathbf{z}^{t} \right) \odot \mathbf{h}^{t{-}1} +\mathbf{z}^{t} \odot \tilde{\mathbf{h}}^{t}
\end{align*}
\end{eqblock}
\\
All convolution operations can be executed in parallel, as they exhibit no temporal dependency via hidden state feedback. The only remaining sequential component is the linear recurrence, which depends solely on the input sequence and the sequence of gate activations. This structure admits efficient computation via parallel scan when operating in log-domain as suggested in \cite{feng2024were}. That is, with $\log(\sigma(x)) = \text{-}\operatorname{Softplus}(\text{-}x)$ we obtain:
\begin{align}
    \log(\hat{\boldsymbol{\upphi}}) &= -\operatorname{Softplus}(\operatorname{Softplus}(-\novelb{\operatorname{Conv}_{\upphi}}[\mathbf{x}]) - \operatorname{Softplus}(-\novelb{\operatorname{Conv}_{\upiota}}[\mathbf{x}]))\\
\log(\hat{\boldsymbol{\upiota}}) &= -\operatorname{Softplus}(\operatorname{Softplus}(-\novelb{\operatorname{Conv}_{\upiota}}[\mathbf{x}]) - \operatorname{Softplus}(-\novelb{\operatorname{Conv}_{\upphi}}[\mathbf{x}]))
\end{align}
Logarithmic normalized gates can be computed time-independently without costly division or sigmoid functions. In the following, we further simplify the gating mechanism.

\subsection{MinConvExpLSTM}
Motivated by the log-domain computations and inspired by \cite{Beck2024xLSTM}, we propose a variant of exponential gating in MinConvLSTM to further improve the model efficiency. To accomplish this, we replace sigmoid by the exponential function:
\begin{align}
    \boldsymbol{\upphi} &= \novelc{\operatorname{exp}}\left( {\operatorname{Conv}_{\upphi}}[\mathbf{x}] \right)\\
    \boldsymbol{\upiota} &= \novelc{\operatorname{exp}}\left( {\operatorname{Conv}_{\upiota}}[\mathbf{x}] \right)
\end{align}
Then, by exploiting the relation between sigmoidal and exponential functions, the normalized forget and input gate take the following form:
\begin{align}
\hat{\boldsymbol{\upphi}}{} &= 
\frac{\boldsymbol{\upphi}}{\boldsymbol{\upphi} + \boldsymbol{\upiota}}=
\sigma\left(
\operatorname{Conv}_{\upphi}[\mathbf{x}] -
\operatorname{Conv}_{\iota}[\mathbf{x}]
\right)\\
\hat{\boldsymbol{\upiota}} &= 
\frac{\boldsymbol{\upiota}}{\boldsymbol{\upphi} + \boldsymbol{\upiota}}=
\sigma\left(
\operatorname{Conv}_{\iota}[\mathbf{x}] -
\operatorname{Conv}_{\upphi}[\mathbf{x}]\right)
 = 1.0 - \hat{\boldsymbol{\upphi}}{}
\end{align}
This formulation enables both gates to be derived from a single sigmoid computation, enhancing computational efficiency. Accordingly, we can basically take the logarithm of these expressions to obtain $\log(\hat{\boldsymbol{\upphi}})$ and $\log(\hat{\boldsymbol{\upiota}})$.

\section{Experiments and Results}

We perform two sets of experiments to benchmark our MinConvGRU and MinConvLSTM architectures against the established ConvGRU and ConvLSTM models.
In the first set of experiments, we train on incompressible Navier-Stokes dynamics in a periodic domain. Second, we investigate the model capacities to learn and predict the real-world weather dynamics dataset, called geopotential. Subsequently, we discuss runtime differences and touch on results produced on tensor processing units (TPUs), where we benchmark PyTorch's convolution operation runtime with tensors of different sizes.

\subsection{Navier-Stokes}
\begin{figure}[t]
    \centering
    \includegraphics[width=\linewidth]{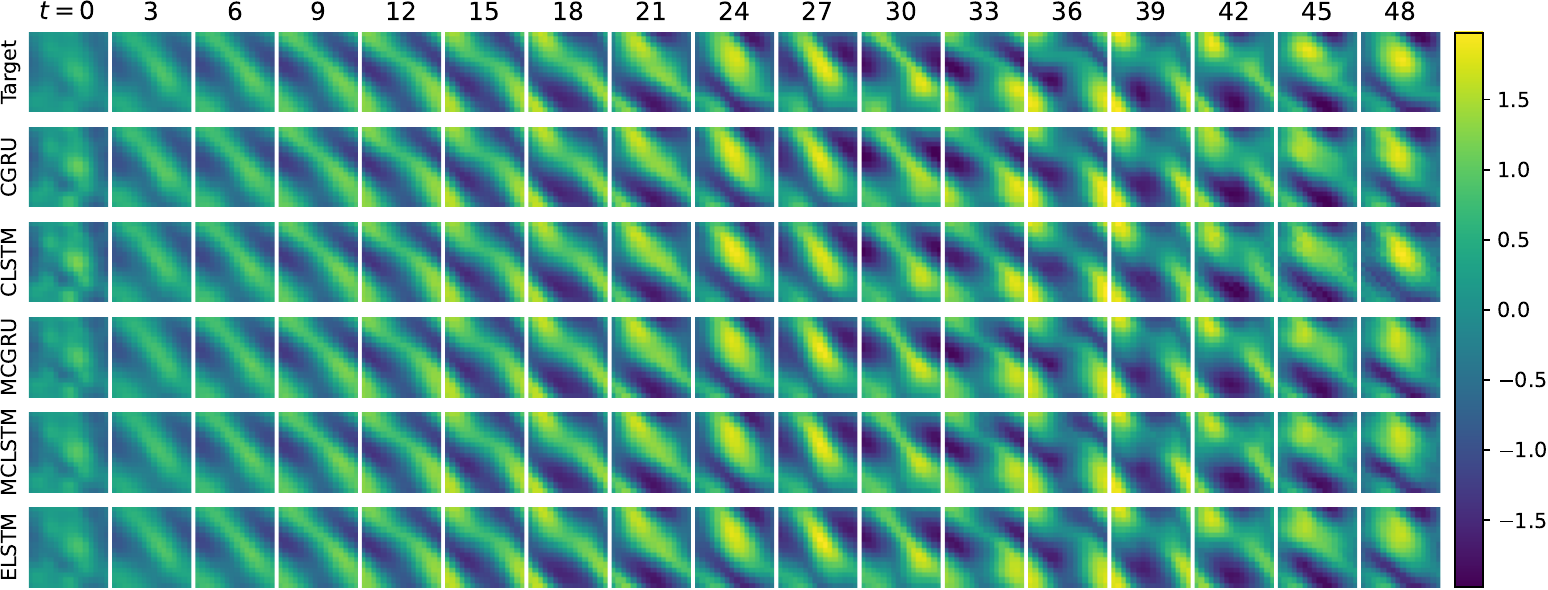}
    \vspace{-0.7cm}
    \caption{Image series of different models (rows) at varying lead time (cols) predicting Navier-Stokes dynamics for 30 time steps in closed loop after 20 steps teacher forcing. First row depicts ground truth, subsequent rows ConvGRU, ConvLSTM, MinConvGRU, MinConvLSTM, and MinConvExpLSTM.}
    \label{fig:ns_image_sequence}
    \vspace{-0.3cm}
\end{figure}
Navier-Stokes equations typically find application in fluid simulation and can be used to generate highly nonlinear dynamics.
Prominent examples include the simulation and prediction of atmospheric weather dynamics, and flow simulations to determine the air resistance in object design.

\compactparagraph{Data Generation} We adhere to \cite{karlbauer2024comparing} and discretize the domain to a $64\times64$ grid when simulating spatiotemporal dynamics over 50 time steps with Reynolds Number $Re=\SI{1e3}{}$ and periodic boundary condition.
To reduce the computational overhead, we further downsample the spatial domain by a factor of four and obtain a resolution of $16\times16$ pixels per time step.
We generate 1000, 50, and 200 samples for training, validation, and testing.

\compactparagraph{Model Configurations} Since each model implements different amounts of weight matrices per unit, we adjust the number of hidden channels per layer such that all models count roughly \SI{175}{k} parameters.
Following a $1\times1$ convolution as encoder to project the input into a latent space with depth $c$, we construct four hidden layers featuring either ConvGRU, ConvLSTM, MinConvGRU, MinConvLSTM, or MinConvExpLSTM cells with $c=28$, $25$, $49$, $40$, and $40$, respectively. Another $1\times1$ convolution eventually projects the evolved latent state back to the data domain with channel depth 1. We apply skip connections, layer and group normalization for further regularization and, in particular, to introduce nonlinearity across linear layers in proposed models.

\compactparagraph{Training Configuration} We use a batch size of one when training all models with Adam and a weight decay of $\SI{1e-2}{}$.
Using a cosine learning rate scheduler, we decay the initial learning rate $\eta_0=\SI{5e-4}{}$ all the way to zero over 30 epochs.
We randomly crop sequences of length 25 from the data and use 20 steps of teacher forcing, followed by five steps of closed loop during training on an NVIDIA RTX 1060 GPU with \SI{6}{GB}.

\begin{figure}[t]
    \centering
    \includegraphics[width=\linewidth]{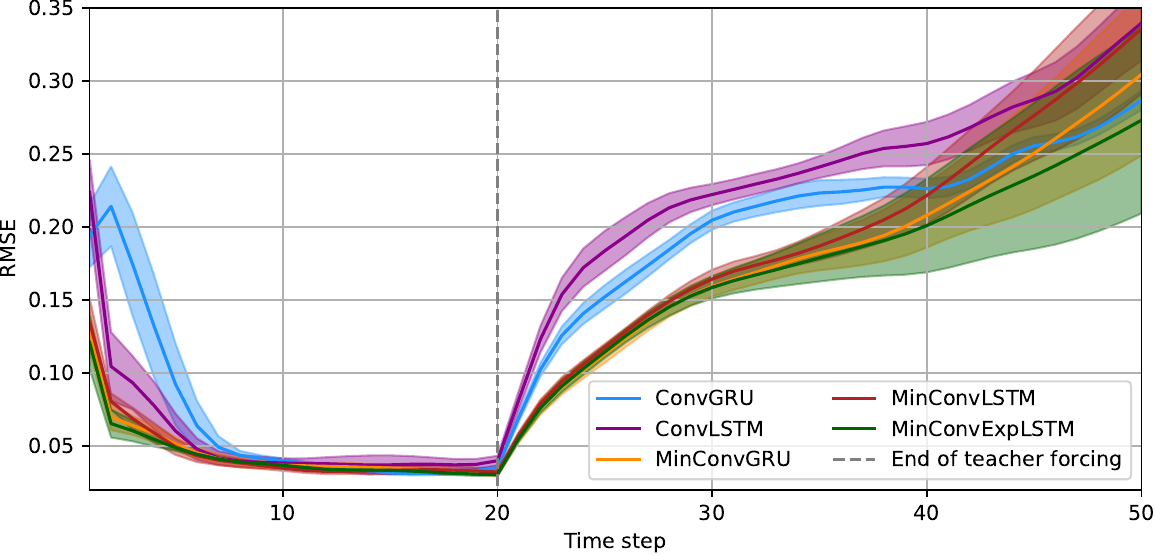}
    \vspace{-0.7cm}
    \caption{
    Root mean squared error of different models over time at predicting Navier-Stokes dynamics. The dashed vertical line at $t=20$ indicates the transition from teacher forcing to closed loop, where models start to autoregressively unroll a prediction into the future. Solid lines and shaded areas indicate the mean and standard deviation over five models trained with different random seeds.}
    \label{fig:ns_rmse_over_time}
\end{figure}

\begin{table}[!t]
    \centering
    \caption{Prediction error comparison of different models on Navier-Stokes and geopotential dynamics, trained on a sequence length of 25 and 24 steps, respectively. 20 steps of teacher forcing is used for both datasets. RMSE$_{\operatorname{TF}}$ and RMSE$_{\operatorname{CL}}$, multiplied by $1\times10^{2}$ for Navier-Stokes and by $1\times10^{-2}$ for geopotential, are computed on a separate test set with sequence length 50 and 20 steps teacher forcing for Navier-Stokes and 20 teacher forcing steps of 96 time steps for geopotential.}
    \vspace{-0.2cm}
    \small
    \begin{tabularx}{\linewidth}{lCCCc}
        \toprule
        & \multicolumn{2}{c}{Geopotential} & \multicolumn{2}{c}{Navier-Stokes}\\
        \cmidrule(r){2-3}\cmidrule(l){4-5}
        Model & RMSE$_{\operatorname{TF}}$ & RMSE$_{\operatorname{CL}}$ & RMSE$_{\operatorname{TF}}$ & RMSE$_{\operatorname{CL}}$\\
        \midrule
        ConvGRU & $1.88\pm0.12$ & $21.52\pm5.18$ & $6.90\pm0.96$ & $20.95\pm0.66$\\
        ConvLSTM & $0.73\pm0.09$ & $12.76\pm2.30$\Bstrut & $5.70\pm0.82$ & $23.81\pm1.32$\Bstrut\\
        \hdashline
        \Tstrut%
        MinConvGRU & $0.88\pm0.23$ & $12.61\pm2.12$ & $4.55\pm0.36$ & $18.57\pm1.98$\\
        MinConvLSTM & $\mathbf{0.70\pm0.08}$ & $\mathbf{10.80\pm1.49}$ & $4.57\pm0.40$ & $19.79\pm1.59$\\
        MinConvExpLSTM &  $0.71\pm0.09$ & $10.94\pm1.37$ & $\mathbf{4.37\pm0.28}$ & $\mathbf{17.87\pm2.48}$\\
        \bottomrule
    \end{tabularx}
    \label{tab:rmses}
    \vspace{-0.4cm}
\end{table}

\compactparagraph{Results} When testing our models, we use 20 steps of teacher forcing followed by 30 steps in closed-loop mode. We qualitatively visualize an exemplary sequence of Navier-Stokes dynamics along with the predicted rollouts of each model in \autoref{fig:ns_image_sequence}.
On first glance, all models manage to predict stable and realistic Navier-Stokes dynamics.
In the last time steps, we find ConvGRU and ConvLSTM deviate from the ground truth by loosing contours (which we attribute to smoothing) as emphasized in frame 44 of \autoref{fig:ns_image_sequence}.
These tendencies are supported quantitatively in \autoref{tab:rmses} as well as in \autoref{fig:ns_rmse_over_time}, which details the RMSE of all models over time on the test set.
Solid lines indicate the average of five independent models trained with different random seeds, whereas the shaded areas mark the respective standard deviations.
We find MinConvGRU, MinConvLSTM, and MinConvExpLSTM outperform their ConvGRU and ConvLSTM counterparts both during teacher forcing and in closed loop prediction.

\begin{figure}[!t]
    \centering
    \includegraphics[width=\linewidth]{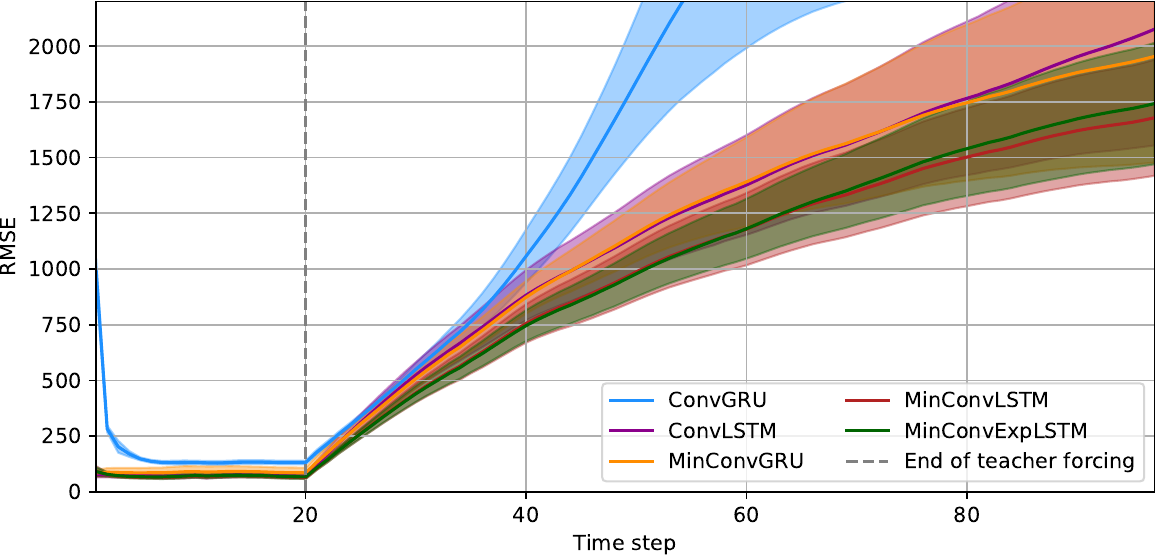}
    \vspace{-0.7cm}
    \caption{
    Root mean squared error of different models over time at predicting global geopotential dynamics. The dashed vertical line at $t=20$ indicates the transition from teacher forcing to closed loop, where models start to autoregressively unroll a prediction into the future. 
    Solid lines and shaded areas indicate the mean and standard deviation over five models trained with different random seeds. The loss values are scaled by $10^{2}$.
    }
    \label{fig:geopot_rmse_over_time}
\end{figure}

\subsection{Geopotential}
As a real-world benchmark, we consider geopotential at \SI{500}{hPa}, denoted as $\Phi_{500}$. This dataset represents the height above sea level (in meters) where the atmospheric pressure is 500 hPa (hectopascals), which equals to an altitude of roughly \SI{5.5}{km} above sea level. It's essentially a way to describe the middle of the atmosphere and is widely used in weather forecasting and climate studies. $\Phi_{500}$ marks a core variable in atmospheric science that characterizes both large scale regional and global meteorological phenomena.

\compactparagraph{Data Preparation} We download all available $\Phi_{500}$ fields from the WeatherBench dataset \cite{rasp2024weatherbench}.
We consider the coarsest resolution of \SI{5.625}{^\circ}, which counts $32\times64$ pixels spanning our Earth, and further downsample the spatial resolution by a factor of two to obtain a resolution of $16\times32$ pixels per hourly time step.
Following common practice, we use the years 1979--2014 for training, 2015--2016 for validation, and 2017--2018 for testing.

\compactparagraph{Model Configurations} In this series of experiments, we follow the parameter configuration as used for learning the Navier-Stokes dynamics with a small modification:
We use three instead of four layers with $c = 14, 12, 24, 20$, and $20$ units (channels) for ConvGRU, ConvLSTM, MinConvGRU, MinConvLSTM, and MinConvExpLSTM, respectively.

\compactparagraph{Training Configuration} We use a batch size of one when training all models with Adam and a weight decay of $\SI{1e-2}{}$.
Using a cosine learning rate scheduler, we decay the initial learning rate $\eta_0=\SI{5e-4}{}$ all the way to zero over 20 epochs.
We randomly crop sequences of length 24 from the data and use 20 steps of teacher forcing, followed by four steps of closed loop during training on an NVIDIA RTX 1060 GPU with \SI{6}{GB}.

\compactparagraph{Results} In line with findings in previous experiments, we find the MinConvRNN variants outperform their traditional counterparts, as visualized in \autoref{fig:geopot_rmse_over_time} and quantified in \autoref{tab:rmses}.

\subsection{Runtime Analysis}
\begin{table}[!t]
    \ra{1.2}
    \centering
    \caption{ Runtime of each model in seconds to train a single epoch in the Navier-Stokes and geopotential experiments. We conducted the same experiments with and without just-in-time compilation (JIT) using \texttt{torch.compile}. The results are averaged over the epochs 2-6. The first epoch is omitted because of the long warm-up phase of \texttt{torch.compile}.}
    \vspace{-0.2cm}
    \small
    \begin{tabularx}{\linewidth}{lCCCc}
        \toprule
        & \multicolumn{2}{c}{Geopotential} & \multicolumn{2}{c}{Navier-Stokes}\\
        \cmidrule(r){2-3}\cmidrule(l){4-5}
        Model & With JIT & No JIT & With JIT & No JIT\\
        \midrule
        ConvGRU & $232.74 \pm 0.88$ & $774.77 \pm 2.46$ & $26.87 \pm 0.074$ & $87.05 \pm 0.162$\\
        ConvLSTM & $182.41 \pm 0.91$ & $513.82 \pm 0.89$\Bstrut & $26.46 \pm 0.034$ & $60.50 \pm 0.317$\Bstrut\\
        \hdashline
        \Tstrut%
        MinConvGRU & $71.944 \pm 0.99$ & $154.97 \pm 1.83$ & $9.62 \pm 0.111$ & $20.33 \pm 0.063$\\
        MinConvLSTM & $74.846 \pm 0.63$ & $170.25 \pm 0.69$ & $8.222 \pm 0.32$ & $19.74 \pm 0.065$\\
        MinConvExpLSTM &  $\mathbf{69.79 \pm 0.855}$ & $\mathbf{154.29 \pm 0.67}$ & $\mathbf{7.956 \pm 0.040}$ & $\mathbf{18.49 \pm 0.015}$\\
        \bottomrule
    \end{tabularx}
    \label{tab:runtimes}
\end{table}

\begin{table}[!t]
    \centering
    \caption{Runtime in ms to process tensors of shape $4\times100\times1\times\operatorname{H}\times\operatorname{W}$ ten times either sequentially over the time dimension (100) or in parallel on a T4 GPU or on a v2-8 TPU in Colab. $\operatorname{H}=\operatorname{W}$ is varied over the table columns.}
    \vspace{-0.2cm}
    \small
    \begin{tabularx}{\linewidth}{p{0.4cm}lRRRRRR}
        \toprule
        & & $4\times4$ & $32\times32$ & $128^2$ & $256^2$ & $512^2$ & $1024^2$\\
        \midrule
        \multirow{3}{*}{\rotatebox{90}{GPU}} & Sequential & 75.56 & 79.34 & 76.01 & 81.20 & 156.95 & 545.70\\
        & Parallel & 0.54 & 0.53 & 7.05 & 27.71 & 112.36 & 465.45\\
        & Speed-up & $\times139$ & $\times149$ & $\times10$ & $\times3$ & $\times1.4$ & $\times1.2$ \Bstrut\\
        \hdashline\Tstrut
        \multirow{3}{*}{\rotatebox{90}{TPU}} & Sequential & 0.109 & 0.243 & 0.257 & 0.206 & 0.185 & 0.427\\
        & Parallel & 0.001 & 0.001 & 0.002 & 0.002 & 0.002 & 0.025\\
        & Speed-up & $\times109$ & $\times243$ & $\times129$ & $\times103$ & $\times93$ & $\times17$\\
        \bottomrule
    \end{tabularx}
    \label{tab:speed_benchmark_colab}
    \vspace{-0.3cm}
\end{table}

\compactparagraph{Training Time} A major motivation of our study is to reduce runtime constraints of convolutional recurrent units.
In \autoref{tab:runtimes}, we report the training time for each model in seconds to conclude a single epoch, averaged over the first five epochs. We report also different results with and without just-in-time compilation (JIT) with \texttt{torch.compile}, making the models much faster. ConvGRU takes longest due to two convolution operations that need to be called in order to complete a single forward pass.
All other models only require a single convolution operation.
In Navier-Stokes experiments, we observe an acceleration by a factor of $3$ with JIT and without JIT around $4.7$. On the other hand, minimal models increase the learning speed remarkably by factor $5$ without using JIT. These findings confirm that our minimal convolutional RNNs models yield substantial computational savings, making them highly attractive for large-scale spatiotemporal learning.

\compactparagraph{Benchmarking Convolution Operations} In our experiments, we observed limitations in the parallel processing of large tensors on GPUs in PyTorch.
We therefore conduct runtime analyses by passing tensors of different shapes through a convolution operation either fully in parallel, or sequentially over the time dimension using a T4 GPU and a v2-8 TPU in Colab.

The results, reported in \autoref{tab:speed_benchmark_colab}, reveal two patterns.
First, the speed gain of parallel over sequential processing on the GPU decreases with growing tensor size and almost levels off for a tensor of shape $1\,024\times1\,024$.
Second, on the TPU, the speed gain of parallel processing is retained even for large tensors of shape $512\times512$, but also decreases beyond.
We observe similar hardware-constrained effects when comparing the sequential ConvRNNs with our MinConvRNNs.

\section{Conclusion}

Processing and learning spatiotemporal dynamics with RNNs has traditionally incurred high runtimes due to the inherently sequential nature of recurrence. In this work, we reformulate the well-established ConvGRU and ConvLSTM architectures by adopting the log-domain scan-compatible structure introduced in MinRNNs \cite{feng2024were}, resulting in MinConvGRU and MinConvLSTM, two minimal convolutional RNNs capable of processing entire spatiotemporal sequences in parallel. Moreover, we integrate exponential gating into the MinConvLSTM to further simplify computation in the log-domain, yielding MinConvExpLSTM.

In two benchmarks, we show that MinConvRNNs are competitive in terms of accuracy while significantly accelerating training. Notably, our minimal models achieved training speed-ups of up to 5x compared to their conventional ConvRNN counterparts, with MinConvExpLSTM attaining the lowest prediction error in the fluid dynamics benchmark.
Subsequent analyses on GPUs and TPUs reveal hardware-related limitations in parallel processing of large tensors. 
On massively parallel hardware, convolution operations may fallback to less efficient implementations when input tensors exceed certain size limits. 

We conclude that MinConvRNNs hold substantial promise for fully parallelized training of spatiotemporal dynamics, enabling the scaling of recurrent architectures in spatiotemporal domains.

Future work could address multi-scale or hierarchical extensions of MinConv architectures to better model spatial dynamics across multiple resolutions. Evaluating these models on higher-resolution or more complex datasets, such as precipitation nowcasting or ocean dynamics, may also reveal scalability limits. 
Additionally, combining minimal recurrent structures with attention mechanisms may offer a promising hybrid approach, balancing efficient scan-compatible dynamic memory with global context modeling.
\bibliographystyle{splncs04}
\bibliography{references}

\end{document}